%% file: main.tex
\documentclass[letterpaper, 10 pt, conference]{ieeeconf}  % Comment this line out if you need a4paper

\IEEEoverridecommandlockouts                              % This command is only needed if 
\usepackage{graphicx} % for pdf, bitmapped graphics files
\usepackage{transparent} % 
\usepackage{times} % assumes new font selection scheme installed
\usepackage{amsmath} % assumes amsmath package installed
\usepackage{amssymb} % assumes amsmath package installed
\usepackage{xcolor}
\usepackage{xspace}
\usepackage{bm}

\usepackage{url}

\newcommand{\ie}{\textit{i.e.}\xspace}

\newcommand{\etc}{\textit{etc.}\xspace}
\newcommand{\cf}{\textit{cf.}\xspace}
\newcommand{\FigRef}[1]{Figure~\ref{#1}}
\newcommand{\TabRef}[1]{Table~\ref{#1}}
\newcommand{\SecRef}[1]{Section~\ref{#1}}
\renewcommand{\vec}{\bm}
\newcommand{\mat}{\mathbf}

\title{\LARGE \bf
Deriving Rewards for Reinforcement Learning from Symbolic Behaviour Descriptions of Bipedal Walking
}

\author{Daniel Harnack$^{1}$, Christoph L\"uth$^{2,3}$, Lukas Gross$^{1}$, Shivesh Kumar$^{1}$, and Frank Kirchner$^{1,3}$
\thanks{This work was supported by the VeryHuman project (grant number 01IW20004) funded the Federal Ministry of Education and Research (BMBF).}% <-this % stops a space
\thanks{$^{1}$ Robotics Innovation Center, DFKI, 28359 Bremen, Germany.}%
\thanks{$^{2}$ Cyber-Physical Systems, DFKI, 28359 Bremen, Germany.}%
\thanks{$^{3}$ University of Bremen, 28359 Bremen, Germany.}%
}

\begin{document}

\maketitle
\thispagestyle{empty}
\pagestyle{empty}

\begin{abstract}
  Generating physical movement behaviours from their symbolic description is a
  long-standing challenge in artificial intelligence (AI) and robotics,
  requiring insights into numerical optimization methods as well as into
  formalizations from symbolic AI and reasoning. In this
  paper, a novel approach to finding a reward function from a symbolic
  description is proposed. The intended system behaviour is modelled as a hybrid
  automaton, which reduces the system state space to allow more
  efficient reinforcement learning. The approach is applied to bipedal walking,
  by modelling the walking robot as a hybrid automaton over state space
  orthants, and used with the compass walker to derive a reward that
  incentivizes following the hybrid automaton cycle. As a result, training times
  of reinforcement learning controllers are reduced while final walking speed
  is increased. The approach can serve as a blueprint how to generate reward
  functions from symbolic AI and reasoning.
\end{abstract}

\section{Introduction}
\label{sec:int}
\input{introduction.tex}

\section{Symbolic Formalization}
\label{sec:sym_form}
\input{symbolic_formalization.tex}

\section{Reward Formulation}
\label{sec:rew_form}
\input{reward_formulation.tex}

\section{Results and Discussion}
\label{sec:res_and_disc}
\input{results_and_discussion.tex}

\addtolength{\textheight}{-20mm}

\section{Conclusion and Outlook}
\label{sec:con_and_out}
\input{conclusion_and_outlook.tex}

\input{bibliography.tex}
\typeout{get arXiv to do 4 passes: Label(s) may have changed. Rerun}

\end{document}

%% file: introduction.tex
Many problems, in particular in robotics, can be phrased as optimization
problems. Popular approaches to solve optimization problems are unsupervised
learning techniques, such as reinforcement learning
(RL)~\cite{sutton2018reinforcement, bertsekas2019reinforcement}. RL requires
both the definition of a \emph{reward function}, which characterizes the quality
of a solution, and an optimization algorithm to reach the maximum reward. When
solving a specific problem arguably most of the time is spent on the former:
the definition of a reward function that captures the essence of the desired
solution, is actually maximized by the desired solution, and allows the
algorithm to follow a gradient towards the desired solution. Poorly designed
reward functions can lead to the algorithm getting stuck in local minima, very
slow initial learning if gradients of the reward are shallow, or solutions that
yield high rewards, but that do not resemble the intended target behaviour.
While reward functions that yield good results for specific behaviours are
discovered over time, there is no principled way of translating a symbolic
description as a human would give it to a numerical reward function that leads
to this behaviour if maximized. The contribution of this paper is a principled
way to derive such reward functions from symbolic descriptions. We tackle the
problem of bipedal walking behaviour, but the general approach can be applied to
many other problems. %  as well.

Walking is a highly relevant locomotion mode in robotics. Whereas there are many
proven combinations of reward formulations and RL algorithms in the literature
for bipedal~\cite{2021_periodic_reward_cassie, rudin2022learning,
li2021reinforcement, yu2018learning, green2021learning} and
quadrupedal~\cite{smith2022walk, rudin2022learning, lee2020learning,
fu2021minimizing} robots, both in simulation and directly on real world
robots~\cite{smith2022walk, tedrake2005learning}, the reward terms are mostly
heuristically generated, and there is no consensus about which reward terms
are necessary or sufficient.
Finding a principled way to infer a reward function with minimal heuristics is
the topic of inverse reinforcement learning (IRL). More precisely, IRL tries to
solve the problem of inferring the reward function of an agent, given its policy
or observed behaviour (see~\cite{ARORA2021103500} for a survey). This may be
complicated, as it requires a very precise understanding of the solution space;
the reward function not only has to characterize the optimal solutions, it also
has to guide the heuristic towards it without creating too many local minima on
the way, as mentioned before. This precise understanding can be difficult to
infer from observations of desired behaviour.

In contrast, humans are able to learn new behaviours from informal verbal or
symbolic descriptions, also without observing demonstrations or teachers as used
in IRL. Such approaches have not yet been well-studied for generating physical
movements in underactuated robots. Thus, in this paper, we try to use symbolic
descriptions of behaviour, as a teacher or coach could give, to derive a
specific reward term for walking.
For this, we use descriptions of the human gait as a succession of certain
phases, such as ``stance foot is in front of swing foot'' or ``swing leg is
moving forward''. At first glance, it is not straightforward to put these
notions into formulas. However, these formal descriptions specify the humanoid's
configuration in certain orthants of its phase space. This observation leads us
to a formal characterization of the human gait given by a succession of phase
space orthants. In the context of RL, we show that this characteristic can be
exploited to effectively reduce the size of the search space when incorporated
in the reward function. Using the compass gait as an example, we show that our
method indeed improves learning times and walking speed. The simplicity of our
technique should make it applicable to other walking robots or similar
situations where a good reward function is hard to come by.

\paragraph*{Organization}
\SecRef{sec:sym_form} presents the formalization of the compass walker gait as a hybrid automaton over
state space orthants. \SecRef{sec:rew_form} uses
this symbolic formalization to derive a reward function term. 
Experimental results are presented in \SecRef{sec:res_and_disc}.
\SecRef{sec:con_and_out} draws the conclusion and discusses future research directions.

%% file: symbolic_formalization.tex
\begin{figure}[t]
  \centering
  \includegraphics[width=\linewidth]{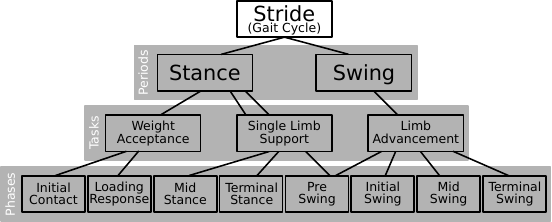}
  \caption{Gait cycle taxonomy according to \cite{perry1992}.}
  \label{fig:gait-cycle}
\end{figure}

\subsection{From Informal to Formal Descriptions}

\begin{figure}[t]
  \centering
  \includegraphics[width=0.9\linewidth]{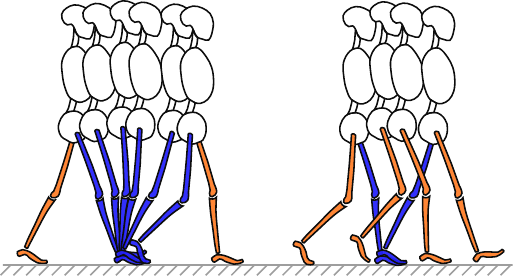}
  \caption{An example of different phases of the human gait cycle, adapted from \cite{perry1992}: the stance leg (blue) supports the upper body (left), while the swing leg (orange) moves freely (right).}
  \label{fig:swing-stance}
\end{figure}

Human walking behaviour is not easy to formalize mathematically. It
consists of various phases serving a different purpose each,
which are combined to an overall walking behaviour. It becomes 
even more complicated when considering legged locomotion 
variations such as running or hopping. 
For example, Perry~\cite{perry1992} decomposes the stride (gait
cycle) into different phases, grouped by tasks according to their
function (\FigRef{fig:gait-cycle}); different tasks accomplish
weight acceptance, limb support and limb advancement (\ie the weight
is supported by one limb while the other swings forward; see \FigRef{fig:swing-stance}).
The individual phases are described informally: ``The swing foot 
lifts until the body weight is aligned over the forefoot 
of the stance leg'', or 
``The second phase begins as the swinging limb is opposite 
of the stance limb. The phase ends when the swinging limb 
is forward and the tibia is vertical, \ie hip and knee flexion 
postures are equal'' (\cite{perry1992} pp. 9--16). 

The aspect to emphasize here is that phases and especially transitions between
two phases are characterized by relations of body parts, some specific joint
angle, or sign changes of angular velocities. Thus, carefully defining the
angles that describe a bipedal system allows us to associate each phase with a
hypercube within the system's phase space. Therefore, we can give a formal
characteristic of a human gait sequence by a set of hypercubes in state space
and a rule in which order to traverse them. The nature of the description lends
itself to a formalization in terms of hybrid automata.

In the following we apply these concepts to the compass gait model, as
a simple and well studied system to test our approach. Despite its
simplicity, it can exhibit a passive walking behaviour on a slope
actuated by gravity~\cite{goswami1996compass, spong1999passivity}.  We
consider an actuated version on flat ground, where a virtual gravity
controller can be used to generate the same gait
pattern~\cite{asano2004novel, asano2005biped}.
We first recall the system dynamics. % , and then define the

\begin{figure}[t]
	\centering
	\includegraphics[width=\linewidth]{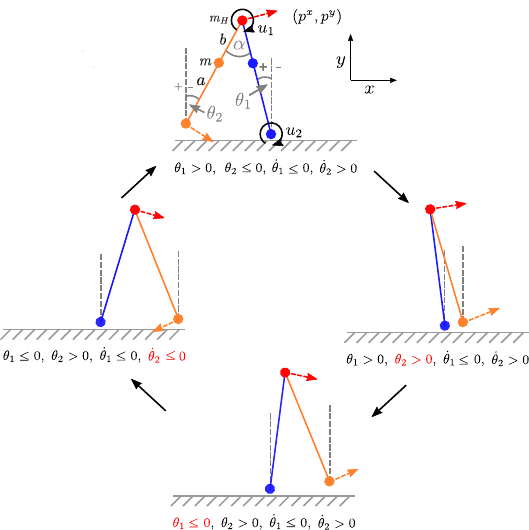}
	\caption{Actuated compass walker cycling through the four orthants sufficient for 
	stable walking. The stance leg is blue, the swing leg is orange. After this cycle, stance and swing leg (and hence $\theta_1, \dot{\theta_1}$ and $\theta_2, \dot{\theta_2}$) are swapped.}
	\label{fig:compass_walker_orthants}
\end{figure}

\subsection{Dynamics of the Compass Walker}

The compass walker~\cite{goswami1996compass, spong1999passivity} is a simple
bipedal walker. Its two legs are called the \emph{stance leg}, which connects with the
ground and supports the weight, and the \emph{swing leg}, which moves freely from the
hip joint. It behaves like a double pendulum with a pin joint at the foot of the
stance leg. Consequently, its dynamics are described as follows:
\begin{align}
\label{eq:walker-1}
\vec{M}(\vec{\theta})\ddot{\vec{\theta}} + \vec{C}(\vec{\theta}, \dot{\vec{\theta}}) + \vec{g}(\vec{\theta}) = \mat{S}\vec{u}
\end{align}
with $\vec{\theta} = [\theta_1, \theta_2]^T$ being the configuration vector,
where $\theta_1$ is the angle of the stance leg to the upright, and $\theta_2$
is the angle between the swing leg and the upright, $\vec u= [u_1, u_2]^T$ the
torque vector, with $u_1$ the torque at the hip and $u_2$ the torque at the
ankle, and $(p^x, p^y)$ the coordinates of the hip joint (\cf{}
\FigRef{fig:compass_walker_orthants}). 

Gravity acts according to 
\begin{equation}
	\vec{g}(\vec{\theta}) = g 
	\begin{bmatrix}
	-(m_H l + ma + ml)\sin(\theta_1) \\ mb\sin(\theta_2)
	\end{bmatrix}
\end{equation}
where $m$ is the mass of leg, given as single mass point at $a$ from
the foot, $l= a+b$ is the length of the leg,  and $m_H$ is the mass
point at the hip (\cf{} \FigRef{fig:compass_walker_orthants}). %
Control acts on the system via
\begin{equation}
\vec{S} = 
\begin{bmatrix}
1 & 1 \\ 0 & -1
\end{bmatrix}
\end{equation}
Inertial and Coriolis matrices are described by 
\begin{align}
	\vec{M}(\vec{\theta}) &= 
	\begin{bmatrix}
	m_Hl^2 + ma^2 + ml^2 & -mbl\cos(\theta_1 - \theta_2) \\ -mbl\cos(\theta_1 - \theta_2) & mb^2
	\end{bmatrix} & \\ 
	\vec{C}(\vec{\theta}, \dot{\vec{\theta}}) &= 
	\begin{bmatrix}
	0 & -mbl\sin(\theta_1 - \theta_2)\dot{\theta}_2 \\ mbl\sin(\theta_1 - \theta_2)\dot{\theta}_1 & 0
	\end{bmatrix}
	\label{eq:walker-2}
\end{align}
Ground collision of the swing leg occurs when it is in front of the
stance leg, the task space velocity of the swing leg is negative in $y$, and $y=0$.\footnote{
Notice that during the forward swing, the swing leg will penetrate or touch the ground due to the simplified model assumptions (\ie no knee joint is included); we disregard this in our simulation.}
When ground contact occurs, swing and stance leg immediately change roles and the angular momentum is transferred assuming a perfectly inelastic collision.
This means $\theta_1$ gets assigned to $\theta_2$ and vice versa, and the transition of angular velocities is calculated via
\begin{align}
	&\, & \mat{T}^{+}(\alpha)\dot{\vec{\theta}}^{+} &= \mat{T}^{-}(\alpha)\dot{\vec{\theta}}^{-} \\
	&\Leftrightarrow & \dot{\vec{\theta}}^{+} &= \mat{T}^{+}(\alpha)^{-1}\mat{T}^{-}(\alpha)\dot{\vec{\theta}}^{-}
	\label{eq:collision_transition}
\end{align}
where $^{+}$/$^{-}$ denote states right after/before collision and 
$\alpha = \theta_1^{-} - \theta_2^{-} = \theta_2^{+} - \theta_1^{+}$ is the 
inter-leg angle at the moment of transition. The transition matrices are given by 
\begin{align*}
	\mat{T}^{+}(\alpha) &= \begin{bmatrix}
	m_Hl^2 + ma^2 + ml(l - b c_\alpha) & mb(b - l c_\alpha) \\ 
	-m b l c_\alpha & mb^2
	\end{bmatrix} \\
	\mat{T}^{-}(\alpha) &= \begin{bmatrix}
	(m_Hl^2 + 2mal) c_\alpha - mab & -mab \\ 
	-mab & 0
	\end{bmatrix}
\end{align*}
where $c_\alpha = \cos(\alpha)$.
This formulation largely follows~\cite{asano2004novel, asano2005biped}, 
further mathematical details can be found in~\cite{goswami1996compass}.

\subsection{A Formal Model of Ideal Walking}
\label{sec:hybrid_automaton_ideal_walking}
Now, the link has to be made to relate the states of these
dynamics with the walking phase descriptions mentioned above.
The crucial observation is that the different phases are determined by
whether the swing leg is to the left or to the right of the stance leg,
whether the stance leg is leaning left or right and whether the legs
are moving clockwise or counter-clockwise around their respective
joints. This can be described in terms of $\theta_1$, $\theta_2$ and
the corresponding angular velocities $\dot{\theta}_1$ and
$\dot{\theta}_2$, which spans the phase space
$\vec{x} = [\theta_1, \theta_2, \dot{\theta}_1, \dot{\theta}_2]^T$ of the
compass walker. \FigRef{fig:compass_walker_orthants} shows this
decomposition, where the swing leg starts from its hindmost position
and swings all the way to the front, when it becomes the stance
leg. 
The case distinction of $\theta_1,\theta_2, \dot{\theta}_1, \dot{\theta}_2$ being smaller or larger than zero
yields sixteen partitions (orthants) of the state space, and the
observation above defines a cycle involving only four of these. 
In the following section, we give a formal model of this ideal walking
behaviour. We then use this model to construct a reward function which
encourages the compass walker to adhere to this formal model; in other words, we reward walking ``properly''.

\newcommand{\Orth}[1]{{\cal O}_{#1}}
\newcommand{\Iff}{\Longleftrightarrow}

\begin{figure}[tb]
  \def\svgwidth{\columnwidth}
  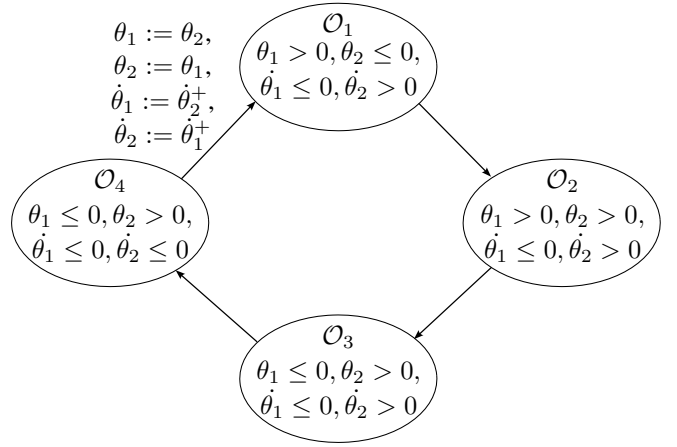
  \caption{Hybrid automaton modelling the orthant sequences. $\dot{\vec{\theta}}^{+} = [\dot{\theta}^{+}_1, \dot{\theta}^{+}_2]^T$
is calculated according to the
impulse transition equation (\ref{eq:collision_transition}).}
  \label{fig:hybrid-automaton}
\end{figure}

The formal model of ideal walking behaviour consists of a cycle of the
four orthants sufficient for stable walking. It is based on hybrid automata, as
introduced by Alur \cite{alur1993} and Henzinger
\cite{henzinger1996a}. Hybrid automata allow combining continuous behaviour with discrete states. Briefly, a hybrid automaton is given as a tuple
\begin{equation}
  {\cal H} = (V_D, Q, E, \mu_1, \mu_2, \mu_3)
\end{equation}
where $V_D$ is a set of $n$ real-valued \emph{data variables} valued in $\Sigma_D= {\mathbb R}^n$, $Q$ is a set of \emph{locations} (discrete states), and $E$ is a \emph{transition relation} $E \subseteq Q\times Q$ between the states.  Further, for each $q\in Q, \sigma\in\Sigma_D$, $\mu_1(q)$ is a set of \emph{activities} specified by \emph{differential equations}, $\mu_2(q, \sigma)$ is an \emph{invariant} predicate, and for each $e\in E$, $\sigma, \tau\in\Sigma_D$ $\mu_3(e, \sigma, \tau)$ is a transition relation between data variables.
Intuitively, the automaton starts in a state $q\in Q$ with data variables set to $\sigma\in \Sigma_D$ s.t. $\mu_2(q, \sigma)$ is true. The continuous behaviour of the data variables is specified by $\mu_1(q)$. As soon as the predicate $\mu_2(q, \sigma)$ becomes false,
the automaton transitions into a state $p\in Q$ where $(q,p)\in E$; the values of the data variables may then change into $\tau\in \Sigma_D$ such that $\mu_3((p, q), \sigma, \tau)$ is true.\footnote{%
Note that although our model is deterministic,  in general hybrid automata can be non-deterministic.}

Here, the state variables are given by
$\theta_1,\theta_2,\dot{\theta}_1,\dot{\theta}_2$,
and we have four locations, corresponding to the states in \FigRef{fig:compass_walker_orthants}:
\begin{align}
  \label{eq:hybrid-1}
  V_D & = \{\theta_1, \theta_2, \dot{\theta}_1,\dot{\theta}_2\} \\
  Q   & = \{\Orth1,\Orth2,\Orth3,\Orth4\}
\end{align}
The activities $\mu_1(q)$ for $q\in Q$ are given by the differential
equations (\ref{eq:walker-1}) to (\ref{eq:walker-2}).
The invariants specify the orthants in the system state as follows:
\begin{align}
  \label{eq:mu-2-a}
  \mu_2(\Orth1) & \Iff \theta_1>0 \land \theta_2\leq 0 \land \dot{\theta_1}\leq 0 \land \dot{\theta_2}>0 \\
  \mu_2(\Orth2) & \Iff \theta_1>0 \land \theta_2 > 0 \land \dot{\theta_1}\leq 0 \land \dot{\theta_2}>0 \\
  \mu_2(\Orth3) & \Iff \theta_1\leq 0 \land \theta_2 > 0 \land \dot{\theta_1}\leq 0 \land \dot{\theta_2}>0 \\
  \mu_2(\Orth4) & \Iff \theta_1\leq 0 \land \theta_2 > 0 \land \dot{\theta_1}\leq 0 \land \dot{\theta_2}\leq 0 
  \label{eq:mu-2-b}
\end{align} 
The transition relation $E$ specifies the cycle mentioned above, and corresponds to the arrows between the four states in \FigRef{fig:compass_walker_orthants}:
\begin{equation}
  E = \{ (\Orth1,\Orth2), (\Orth2,\Orth3), (\Orth3,\Orth4), (\Orth4,\Orth1) \} 
\end{equation}
When a transition occurs, $\mu_3$ specifies the possible values of
$\theta_1, \theta_2, \dot{\theta}_1, \dot{\theta}_2$. This mapping is
the identity, except at the last transition $(\Orth4,\Orth1)$, where
$\theta_1$ and $\theta_2$ are interchanged, and the impulse is transferred
between $\dot{\theta}_1$ and $\dot{\theta}_2$:
\begin{equation}
  \mu_3(p,q) = \left\{
    \begin{array}{ll}
     \{(\theta_1,\theta_2), (\theta_2,\theta_1), (\dot{\theta}_1,\dot{\theta}^{+}_2), (\dot{\theta}_2,\dot{\theta}^{+}_1)\} & \\
      \quad \quad \mbox{if} \, p = \Orth4, q= \Orth1 & \\
     \{(\theta_1,\theta_1), (\theta_2,\theta_2), (\dot{\theta}_1,\dot{\theta}_1), (\dot{\theta}_2,\dot{\theta}_2)\} & \\
     \quad \quad \mbox{otherwise} & \\
    \end{array}\right.
\label{eq:hybrid-2}
\end{equation}
where $\dot{\vec{\theta}}^{+} = [\dot{\theta}^{+}_1, \dot{\theta}^{+}_2]^T$
is calculated according to the
impulse transition equation (\ref{eq:collision_transition}).
The automaton defined by equations (\ref{eq:hybrid-1}) to
(\ref{eq:hybrid-2}) can also be described by a diagram (\cf
\FigRef{fig:hybrid-automaton}). The diagram shows the states, the
state invariants defined by $\mu_2$, the state transitions in $E$, and the
change of the state variables after the transitions defined by
$\mu_3$; it does not show the identities in (\ref{eq:hybrid-2}), \ie
$\theta_1 := \theta_1, \theta_2:= \theta_2$, \etc in all other transitions,
as this is usually elided, and for clarity we also do not show the activities
(defined by $\mu_1$) in the diagram.

%% file: hybrid-automaton.pdf_tex
%% Creator: Inkscape 1.2.1 (9c6d41e, 2022-07-14), www.inkscape.org
%% Accompanies image file 'hybrid-automaton.pdf' (pdf, eps, ps)
\begingroup%
  \makeatletter%
  \providecommand\color[2][]{%
    \errmessage{(Inkscape) Color is used for the text in Inkscape, but the package 'color.sty' is not loaded}%
    \renewcommand\color[2][]{}%
  }%
  \providecommand\transparent[1]{%
    \errmessage{(Inkscape) Transparency is used (non-zero) for the text in Inkscape, but the package 'transparent.sty' is not loaded}%
    \renewcommand\transparent[1]{}%
  }%
  \providecommand\rotatebox[2]{#2}%
  \newcommand*\fsize{\dimexpr\f@size pt\relax}%
  \newcommand*\lineheight[1]{\fontsize{\fsize}{#1\fsize}\selectfont}%
  \ifx\svgwidth\undefined%
    \setlength{\unitlength}{289.18165312bp}%
    \ifx\svgscale\undefined%
      \relax%
    \else%
      \setlength{\unitlength}{\unitlength * \real{\svgscale}}%
    \fi%
  \else%
    \setlength{\unitlength}{\svgwidth}%
  \fi%
  \global\let\svgwidth\undefined%
  \global\let\svgscale\undefined%
  \makeatother%
  \begin{picture}(1,0.6780968)%
    \lineheight{1}%
    \setlength\tabcolsep{0pt}%
    \put(0,0){\includegraphics[width=\unitlength,page=1]{hybrid-automaton.pdf}}%
    \put(0.50410534,0.63322475){\color[rgb]{0,0,0}\makebox(0,0)[t]{\lineheight{1.25}\smash{\begin{tabular}[t]{c}$\Orth1$\end{tabular}}}}%
    \put(0.50420665,0.5898662){\color[rgb]{0,0,0}\makebox(0,0)[t]{\lineheight{1.25}\smash{\begin{tabular}[t]{c}$\theta_1>0, \theta_2\leq0,$\\$\dot{\theta_1}\leq0, \dot{\theta_2}>0$\end{tabular}}}}%
    \put(0,0){\includegraphics[width=\unitlength,page=2]{hybrid-automaton.pdf}}%
    \put(0.50410534,0.15304499){\color[rgb]{0,0,0}\makebox(0,0)[t]{\lineheight{1.25}\smash{\begin{tabular}[t]{c}$\Orth3$\end{tabular}}}}%
    \put(0.50420665,0.09820683){\color[rgb]{0,0,0}\makebox(0,0)[t]{\lineheight{1.25}\smash{\begin{tabular}[t]{c}$\theta_1\leq0, \theta_2>0,$\\$\dot{\theta_1}\leq0, \dot{\theta_2}>0$\end{tabular}}}}%
    \put(0,0){\includegraphics[width=\unitlength,page=3]{hybrid-automaton.pdf}}%
    \put(0.15286224,0.39313489){\color[rgb]{0,0,0}\makebox(0,0)[t]{\lineheight{1.25}\smash{\begin{tabular}[t]{c}$\Orth4$\end{tabular}}}}%
    \put(0.15296354,0.33829672){\color[rgb]{0,0,0}\makebox(0,0)[t]{\lineheight{1.25}\smash{\begin{tabular}[t]{c}$\theta_1\leq0, \theta_2>0,$\\$\dot{\theta_1}\leq0, \dot{\theta_2}\leq0$\end{tabular}}}}%
    \put(0,0){\includegraphics[width=\unitlength,page=4]{hybrid-automaton.pdf}}%
    \put(0.84922382,0.39313489){\color[rgb]{0,0,0}\makebox(0,0)[t]{\lineheight{1.25}\smash{\begin{tabular}[t]{c}$\Orth2$\end{tabular}}}}%
    \put(0.84932516,0.33829672){\color[rgb]{0,0,0}\makebox(0,0)[t]{\lineheight{1.25}\smash{\begin{tabular}[t]{c}$\theta_1>0, \theta_2>0,$\\$\dot{\theta_1}\leq0, \dot{\theta_2}>0$\end{tabular}}}}%
    \put(0.23351524,0.61873551){\color[rgb]{0,0,0}\makebox(0,0)[t]{\lineheight{1.25}\smash{\begin{tabular}[t]{c}$\theta_1:= \theta_2$,\\$\theta_2 := \theta_1$,\\$\dot{\theta}_1:= \dot{\theta}_2^+$,\\$\dot{\theta}_2:= \dot{\theta}_1^+$\end{tabular}}}}%
  \end{picture}%
\endgroup%

%% file: reward_formulation.tex
\begin{figure}[t]
	\centering
	\includegraphics[width=0.95\linewidth]{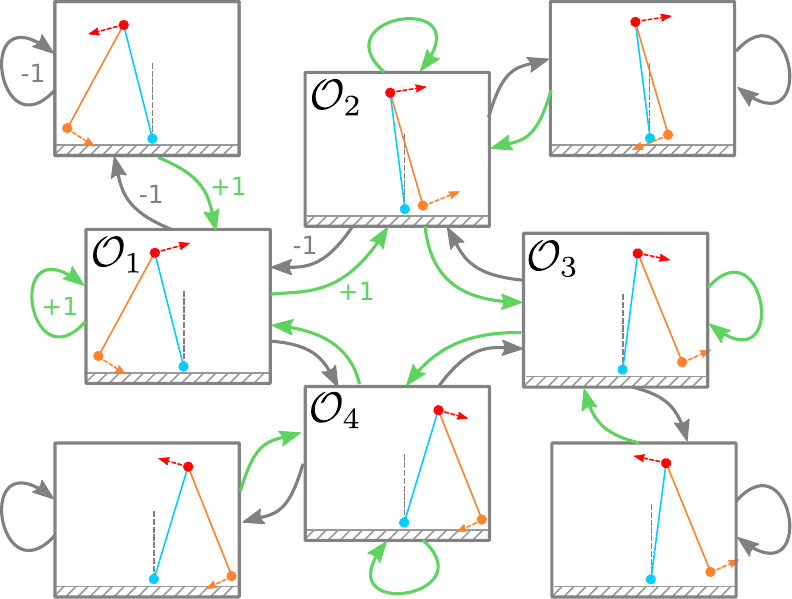}
	\caption{Visualization of the orthant reward term defined in equations (\ref{eq:r_or}).
	Staying in the stable orthant cycle or entering the cycle is rewarded with +1, 
	exiting the cycle or moving in the wrong direction is punished with -1. Not all 
	possible orthants are shown for clarity.}
	\label{fig:orthant_reward}
\end{figure}

For an agent to make use of this exploration space reduction given by the hybrid
automaton formulation, we define the orthant reward term $r_{\mathrm{or}}$ as:
\begin{equation}
	\label{eq:r_or}
	r_{\mathrm{or}}(\vec{x}_{t}, \vec{x}_{t-1}) = \begin{cases}
	+1 \quad \mathrm{if} \, &\mathcal{O}(\vec{x}_{t-1}) \in Q \, \land \\
	&\mathcal{O}(\vec{x}_{t}) \in Q \, \land \\ 
	&(\mathcal{O}(\vec{x}_{t-1}), \mathcal{O}(\vec{x}_{t})) \in E \\
	+1 \quad \mathrm{if} \, &\mathcal{O}(\vec{x}_{t-1}) \notin Q \, \land \\
	&\mathcal{O}(\vec{x}_{t}) \in Q \\
	-1 \quad \mathrm{else}
	\end{cases}
\end{equation}
Here, $Q, E$ refer to the definitions of the hybrid automaton of ideal
walking from \SecRef{sec:hybrid_automaton_ideal_walking}, and $\mathcal{O}(\vec{x})$
maps the state $\vec{x}$ to the current orthant $\mathcal{O}_k$, $k \in \{1, ..., 16\}$.
This reward is visualized in \FigRef{fig:orthant_reward}; essentially, it incentivizes the walker to follow the sequence of the formal model developed in \SecRef{sec:hybrid_automaton_ideal_walking}.

As comparison, we also consider a reward which is commonly used in 
locomotion tasks, namely rewarding task space motion in the intended direction.
This reward reads 
\begin{equation}
	r_{\mathrm{for}}(\vec{p}_t, \vec{p}_{t-1}) = 2H(p_t^{x} - p_{t-1}^{x}) - 1
\end{equation}
where $H$ denotes the Heaviside function, and 
$\vec{p}_t = (p_t^{x}, p_t^{y})$ the task space coordinates of the 
hip joint (see \FigRef{fig:compass_walker_orthants}). 
Further, rewards are used to encourage smooth controls ($r_{\mathrm{jerk}}$), high 
walking distance ($r_{\mathrm{dist}}$), and punish falling ($r_{\mathrm{fall}}$). 
These terms are defined as:
\begin{align}
	r_{\mathrm{jerk}}(\vec{u}_t, \vec{u}_{t-1}) &= ||\vec{u}_t-\vec{u}_{t-1}||_2 \\
	r_{\mathrm{dist}}(\vec{p}_t, t) &= p_t^{x}H(t - T) x\\
	r_{\mathrm{fall}}(\vec{p}_t) &= H(-p_t^{y}) 
\end{align}
The full reward is then given as 
\begin{align}
	\label{eq:rew_formula}
	&r(\vec{x}_t, \vec{x}_{t-1}, \vec{p}_t, \vec{p}_{t-1}, \vec{u}_t, \vec{u}_{t-1}, t) = \\
	&\omega_{\mathrm{jerk}}r_{\mathrm{jerk}}(\vec{u}_t, \vec{u}_{t-1}) + \omega_{\mathrm{dist}}r_{\mathrm{dist}}(\vec{p}_t, t) + \notag \\ 
	&\omega_{\mathrm{fall}}r_{\mathrm{fall}}(\vec{p}_t) +  \omega_{\mathrm{for}}r_{\mathrm{for}}(\vec{p}_t, \vec{p}_{t-1}) + \omega_{\mathrm{or}}r_{\mathrm{or}}(\vec{x}_{t}, \vec{x}_{t-1}) \notag
\end{align}

%% file: results_and_discussion.tex
Using the previously mentioned formulation, 
we compared training setups of reward configurations in terms of 
reward optimization and highest achieved walking distance.\footnote{% 
Accompanying code is publicly available at \url{https://github.com/dfki-ric-underactuated-lab/orthant_rewards_biped_rl}.} %
The observation in the RL setup was the robot configuration $\vec{x}_t$. Thus, a
policy $\pi(\vec{x}_t) \to \vec u_t$ was trained mapping the configuration state
to the control input. As training algorithm, PPO was
chosen~\cite{schulman2017proximal} in the stable-baselines3
implementation~\cite{stable-baselines3} with default parameters. We evaluate
different combinations of reward terms to disentangle the effects of the various
terms. The different reward setups are detailed in \TabRef{tab:rew_setup}. 
\begin{table}[h]
	\centering
	\scriptsize
	\caption{Weights for the reward terms}
	\begin{center}
		\begin{tabular}{c|c|c|c|c}
			& sparse & for & or & for + or \\
			\hline
			$\omega_{\text{for}}$ & 0.0 & 0.01 & 0.0 & 0.005 \\
			\hline
			$\omega_{\text{or}}$ & 0.0 & 0.0 & 0.01 & 0.005 \\
		\end{tabular}
	\end{center}
	\label{tab:rew_setup}
\end{table}
The other weights in equation (\ref{eq:rew_formula}) were set to
$\omega_{\mathrm{jerk}} = -0.001, \omega_{\mathrm{dist}} =1, \omega_{\mathrm{fall}} = -10$. 
A training episode was terminated if a fall occurred, \ie if $r_{\mathrm{fall}}(\vec{p}_t) = 1$, or the maximal episode length of $T=10$~s was reached. Each training condition was run 15 times with different random seeds and 500000 environment steps per run.

As a baseline comparison, we use the virtual gravity controller 
\begin{equation}
	\label{eq:virtual_gravity}
	\mat{S}\vec{u}_{vg} = \begin{bmatrix}
	(m_Hl + m(a + l))\cos(\theta_1) \\
	-mb\cos(\theta_2)
	\end{bmatrix}
	g\tan(\phi)
\end{equation}
described in~\cite{asano2005biped}, simulating passive walking on a slope of $\phi=-0.07$~rad. 
The parameters of the compass walker plant used in all simulations is given in~\TabRef{tab:sim_params}, and the initial condition for all experiments 
was $\vec{x}_{0} = [0.0, 0.0, -0.4, 2.0]^T$.
\begin{table}[h]
	\centering
	\scriptsize
	\caption{Physical Parameters of the Compass Gait Walker}
	\begin{center}
		\begin{tabular}{c|c|c|c|c}
			$m_H$ & $m$ & $a$ & $b$ & $l$\\
			\hline
			1~kg & 0.5~kg & 0.5~m & 0.5~m & 1.0~m \\
		\end{tabular}
	\end{center}
	\label{tab:sim_params}
\end{table}

\begin{figure}[t]
	\centering
	\def\svgwidth{0.95\linewidth}
	\scriptsize	
    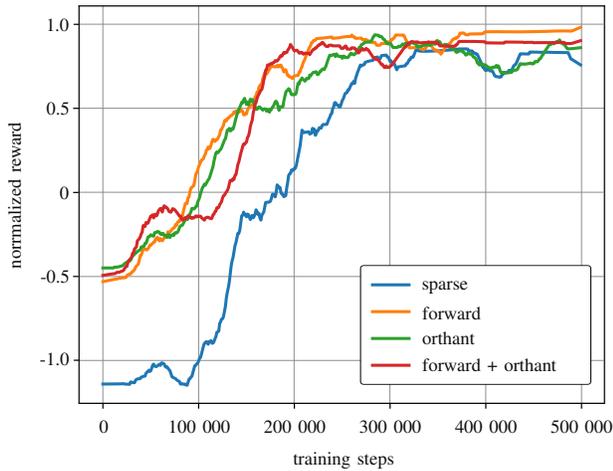
	\caption{Averaged learning curves for different rewards. The virtual gravity
	controller from equation (\ref{eq:virtual_gravity}) serves as baseline to which the
	rewards are normalized.}
	\label{fig:learning_curves}
\end{figure}

\begin{figure}[t]
	\centering
	\def\svgwidth{0.95\linewidth}
	\scriptsize	
    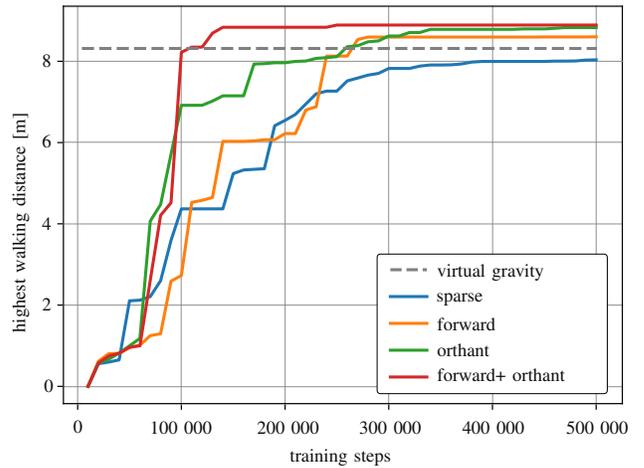
	\caption{Highest walking distances achieved for $t= 10s$ for different
	rewards. The virtual gravity	controller from equation
	(\ref{eq:virtual_gravity}) serves as baseline. All reward setups result in
	higher walking distances than the reference, except for the sparse setup. A
	combination of forward and orthant reward terms yields the highest
	improvements.}
	\label{fig:walking_distance}
\end{figure}

\FigRef{fig:learning_curves} shows the averaged learning curves for the
different reward setups, normalized to the reward gained by the virtual gravity
baseline controller. Whereas reward setups using forward or orthant reward terms
reach close to optimal reward values earlier in training, after around 200000
steps, convergence is slower for the sparse setup. \FigRef{fig:walking_distance}
compares the highest achieved walking distance for each reward setup, taken over
the 15 runs per setup.\footnote{See accompanying video at
\url{https://youtu.be/CkvLvz_tLtc}.} Here, the combined reward of orthant reward
and forward velocity reward leads to the fastest convergence and highest
achieved walking distance. Similar walking distances are reached after
considerably longer time in both individual reward setups of only orthant reward
and only forward velocity reward. Strikingly, walking distances for the sparse
setup also converge, but at markedly lower values. The standard deviation of the
highest achieved reward highest achieved walking distance are shown in
Table~\ref{tab:std_dev}.

\begin{table}[t]
	\centering
	\caption{Standard deviations}
	\begin{center}
		\begin{tabular}{r|c|c|c|c}
		setup &	sparse & forward & orthant & fwd+ orthant \\ \hline
		reward	& $0.127$ & $0.045$ & $0.039$ & $0.088$ \\
		walking distance & 1.302 & 1.192 & 0.744 & 1.173 \\
		\end{tabular}
	\end{center}
	\label{tab:std_dev}
\end{table}

In summary, whereas a sparse reward setup reaches comparable performance to the
baseline virtual gravity controller, both the typically used forward velocity
reward and the novel orthant reward improve the performance, with the
combination of both reaching the highest performance. The orthant reward
leads to the lowest standard deviation of the performance across trials,
making it the most reproducible method in our study. The combination of the
orthant and forward reward has an increased standard deviation, however below the
values for the sparse reward. Interestingly, both forward
and orthant rewards
seem redundant: Observing a stable compass gait, both the predicates of
'forward hip velocity is positive' and 'orthant sequence follows the stable
orthant gait cycle' are always true. Yet the combination of both incentives
through the reward function is more successful in reaching higher walking distances
than each incentive on its own, at the cost of a slightly reduced reproducibility indicated
by the increased standard deviation.
This is part of a long-standing question in reward shaping, where the optimal
combination of reward terms to achieve a goal is unclear. Too many redundant
terms can deteriorate learning, whereas too little guidance can lead to
impractical training times or no convergence at all. In comparison to typically
used reward terms, the orthant reward is simple and straight forward since it
amounts to a single truth value telling the agent whether it is close to the set
of desired phase space trajectories.

%% file: learning_curves.pdf_tex
%% Creator: Inkscape 1.2.1 (9c6d41e, 2022-07-14), www.inkscape.org
%% Accompanies image file 'learning_curves.pdf' (pdf, eps, ps)
\begingroup%
  \makeatletter%
  \providecommand\color[2][]{%
    \errmessage{(Inkscape) Color is used for the text in Inkscape, but the package 'color.sty' is not loaded}%
    \renewcommand\color[2][]{}%
  }%
  \providecommand\transparent[1]{%
    \errmessage{(Inkscape) Transparency is used (non-zero) for the text in Inkscape, but the package 'transparent.sty' is not loaded}%
    \renewcommand\transparent[1]{}%
  }%
  \providecommand\rotatebox[2]{#2}%
  \newcommand*\fsize{\dimexpr\f@size pt\relax}%
  \newcommand*\lineheight[1]{\fontsize{\fsize}{#1\fsize}\selectfont}%
  \ifx\svgwidth\undefined%
    \setlength{\unitlength}{405.29223633bp}%
    \ifx\svgscale\undefined%
      \relax%
    \else%
      \setlength{\unitlength}{\unitlength * \real{\svgscale}}%
    \fi%
  \else%
    \setlength{\unitlength}{\svgwidth}%
  \fi%
  \global\let\svgwidth\undefined%
  \global\let\svgscale\undefined%
  \makeatother%
  \begin{picture}(1,0.76482689)%
    \lineheight{1}%
    \setlength\tabcolsep{0pt}%
    \put(0,0){\includegraphics[width=\unitlength,page=1]{learning_curves.pdf}}%
    \put(0.07531809,0.71237143){\color[rgb]{0.00392157,0.00392157,0.00392157}\transparent{0.952941}\makebox(0,0)[lt]{\lineheight{1.25}\smash{\begin{tabular}[t]{l}1.0\end{tabular}}}}%
    \put(0.07604095,0.57607634){\color[rgb]{0.00392157,0.00392157,0.00392157}\transparent{0.952941}\makebox(0,0)[lt]{\lineheight{1.25}\smash{\begin{tabular}[t]{l}0.5\end{tabular}}}}%
    \put(0.06618116,0.30397325){\color[rgb]{0.00392157,0.00392157,0.00392157}\transparent{0.952941}\makebox(0,0)[lt]{\lineheight{1.25}\smash{\begin{tabular}[t]{l}-0.5\end{tabular}}}}%
    \put(0.0654583,0.17000801){\color[rgb]{0.00392157,0.00392157,0.00392157}\transparent{0.952941}\makebox(0,0)[lt]{\lineheight{1.25}\smash{\begin{tabular}[t]{l}-1.0\end{tabular}}}}%
    \put(0.09752429,0.43978122){\color[rgb]{0.00392157,0.00392157,0.00392157}\transparent{0.952941}\makebox(0,0)[lt]{\lineheight{1.25}\smash{\begin{tabular}[t]{l}0\end{tabular}}}}%
    \put(0.03490827,0.32720022){\color[rgb]{0.00392157,0.00392157,0.00392157}\transparent{0.952941}\rotatebox{90}{\makebox(0,0)[lt]{\lineheight{1.25}\smash{\begin{tabular}[t]{l}norrmalized reward\end{tabular}}}}}%
    \put(0.55765807,0.00865243){\color[rgb]{0.00392157,0.00392157,0.00392157}\transparent{0.952941}\makebox(0,0)[t]{\lineheight{1.25}\smash{\begin{tabular}[t]{c}training steps\end{tabular}}}}%
    \put(0.1688625,0.06089812){\color[rgb]{0.00392157,0.00392157,0.00392157}\transparent{0.952941}\makebox(0,0)[t]{\lineheight{1.25}\smash{\begin{tabular}[t]{c}0\end{tabular}}}}%
    \put(0.32372179,0.06089812){\color[rgb]{0.00392157,0.00392157,0.00392157}\transparent{0.952941}\makebox(0,0)[t]{\lineheight{1.25}\smash{\begin{tabular}[t]{c}100 000\end{tabular}}}}%
    \put(0.47858102,0.06089812){\color[rgb]{0.00392157,0.00392157,0.00392157}\transparent{0.952941}\makebox(0,0)[t]{\lineheight{1.25}\smash{\begin{tabular}[t]{c}200 000\end{tabular}}}}%
    \put(0.63344028,0.06089812){\color[rgb]{0.00392157,0.00392157,0.00392157}\transparent{0.952941}\makebox(0,0)[t]{\lineheight{1.25}\smash{\begin{tabular}[t]{c}300 000\end{tabular}}}}%
    \put(0.78829959,0.06089812){\color[rgb]{0.00392157,0.00392157,0.00392157}\transparent{0.952941}\makebox(0,0)[t]{\lineheight{1.25}\smash{\begin{tabular}[t]{c}400 000\end{tabular}}}}%
    \put(0.94315882,0.06094149){\color[rgb]{0.00392157,0.00392157,0.00392157}\transparent{0.952941}\makebox(0,0)[t]{\lineheight{1.25}\smash{\begin{tabular}[t]{c}500 000\end{tabular}}}}%
    \put(0.68435396,0.29239525){\color[rgb]{0.00392157,0.00392157,0.00392157}\transparent{0.952941}\makebox(0,0)[lt]{\lineheight{1.25}\smash{\begin{tabular}[t]{l}sparse\end{tabular}}}}%
    \put(0.68527922,0.24461953){\color[rgb]{0.00392157,0.00392157,0.00392157}\transparent{0.952941}\makebox(0,0)[lt]{\lineheight{1.25}\smash{\begin{tabular}[t]{l}forward\end{tabular}}}}%
    \put(0.68527922,0.16060328){\color[rgb]{0.00392157,0.00392157,0.00392157}\transparent{0.952941}\makebox(0,0)[lt]{\lineheight{1.25}\smash{\begin{tabular}[t]{l}forward + orthant\end{tabular}}}}%
    \put(0.68510573,0.20343353){\color[rgb]{0.00392157,0.00392157,0.00392157}\transparent{0.952941}\makebox(0,0)[lt]{\lineheight{1.25}\smash{\begin{tabular}[t]{l}orthant\end{tabular}}}}%
  \end{picture}%
\endgroup%

%% file: walking_distance.pdf_tex
%% Creator: Inkscape 1.2.1 (9c6d41e, 2022-07-14), www.inkscape.org
%% Accompanies image file 'walking_distance.pdf' (pdf, eps, ps)
\begingroup%
  \makeatletter%
  \providecommand\color[2][]{%
    \errmessage{(Inkscape) Color is used for the text in Inkscape, but the package 'color.sty' is not loaded}%
    \renewcommand\color[2][]{}%
  }%
  \providecommand\transparent[1]{%
    \errmessage{(Inkscape) Transparency is used (non-zero) for the text in Inkscape, but the package 'transparent.sty' is not loaded}%
    \renewcommand\transparent[1]{}%
  }%
  \providecommand\rotatebox[2]{#2}%
  \newcommand*\fsize{\dimexpr\f@size pt\relax}%
  \newcommand*\lineheight[1]{\fontsize{\fsize}{#1\fsize}\selectfont}%
  \ifx\svgwidth\undefined%
    \setlength{\unitlength}{405.22476196bp}%
    \ifx\svgscale\undefined%
      \relax%
    \else%
      \setlength{\unitlength}{\unitlength * \real{\svgscale}}%
    \fi%
  \else%
    \setlength{\unitlength}{\svgwidth}%
  \fi%
  \global\let\svgwidth\undefined%
  \global\let\svgscale\undefined%
  \makeatother%
  \begin{picture}(1,0.7551937)%
    \lineheight{1}%
    \setlength\tabcolsep{0pt}%
    \put(0,0){\includegraphics[width=\unitlength,page=1]{walking_distance.pdf}}%
    \put(0.05901552,0.65129468){\color[rgb]{0.00392157,0.00392157,0.00392157}\transparent{0.952941}\makebox(0,0)[t]{\lineheight{1.25}\smash{\begin{tabular}[t]{c}8\end{tabular}}}}%
    \put(0.05890708,0.52064985){\color[rgb]{0.00392157,0.00392157,0.00392157}\transparent{0.952941}\makebox(0,0)[t]{\lineheight{1.25}\smash{\begin{tabular}[t]{c}6\end{tabular}}}}%
    \put(0.05918181,0.38670216){\color[rgb]{0.00392157,0.00392157,0.00392157}\transparent{0.952941}\makebox(0,0)[t]{\lineheight{1.25}\smash{\begin{tabular}[t]{c}4\end{tabular}}}}%
    \put(0.05887816,0.25685494){\color[rgb]{0.00392157,0.00392157,0.00392157}\transparent{0.952941}\makebox(0,0)[t]{\lineheight{1.25}\smash{\begin{tabular}[t]{c}2\end{tabular}}}}%
    \put(0.06223012,0.124274){\color[rgb]{0.00392157,0.00392157,0.00392157}\transparent{0.952941}\makebox(0,0)[t]{\lineheight{1.25}\smash{\begin{tabular}[t]{c}0\end{tabular}}}}%
    \put(0.02423254,0.39891443){\color[rgb]{0.00392157,0.00392157,0.00392157}\transparent{0.952941}\rotatebox{90}{\makebox(0,0)[t]{\lineheight{1.25}\smash{\begin{tabular}[t]{c}highest walking distance [m]\end{tabular}}}}}%
    \put(0.69183077,0.31152743){\color[rgb]{0.00392157,0.00392157,0.00392157}\transparent{0.952941}\makebox(0,0)[lt]{\lineheight{1.25}\smash{\begin{tabular}[t]{l}virtual gravity\end{tabular}}}}%
    \put(0.69086198,0.27092821){\color[rgb]{0.00392157,0.00392157,0.00392157}\transparent{0.952941}\makebox(0,0)[lt]{\lineheight{1.25}\smash{\begin{tabular}[t]{l}sparse\end{tabular}}}}%
    \put(0.69178739,0.22385106){\color[rgb]{0.00392157,0.00392157,0.00392157}\transparent{0.952941}\makebox(0,0)[lt]{\lineheight{1.25}\smash{\begin{tabular}[t]{l}forward\end{tabular}}}}%
    \put(0.69178739,0.14172715){\color[rgb]{0.00392157,0.00392157,0.00392157}\transparent{0.952941}\makebox(0,0)[lt]{\lineheight{1.25}\smash{\begin{tabular}[t]{l}forward+ orthant\end{tabular}}}}%
    \put(0.69161388,0.18280354){\color[rgb]{0.00392157,0.00392157,0.00392157}\transparent{0.952941}\makebox(0,0)[lt]{\lineheight{1.25}\smash{\begin{tabular}[t]{l}orthant\end{tabular}}}}%
    \put(0.10186771,0.05722785){\color[rgb]{0.00392157,0.00392157,0.00392157}\transparent{0.952941}\makebox(0,0)[lt]{\lineheight{1.25}\smash{\begin{tabular}[t]{l}0\end{tabular}}}}%
    \put(0.22624772,0.05593334){\color[rgb]{0.00392157,0.00392157,0.00392157}\transparent{0.952941}\makebox(0,0)[lt]{\lineheight{1.25}\smash{\begin{tabular}[t]{l}100 000\end{tabular}}}}%
    \put(0.39983298,0.05593334){\color[rgb]{0.00392157,0.00392157,0.00392157}\transparent{0.952941}\makebox(0,0)[lt]{\lineheight{1.25}\smash{\begin{tabular}[t]{l}200 000\end{tabular}}}}%
    \put(0.56556154,0.05593334){\color[rgb]{0.00392157,0.00392157,0.00392157}\transparent{0.952941}\makebox(0,0)[lt]{\lineheight{1.25}\smash{\begin{tabular}[t]{l}300 000\end{tabular}}}}%
    \put(0.73231274,0.05593334){\color[rgb]{0.00392157,0.00392157,0.00392157}\transparent{0.952941}\makebox(0,0)[lt]{\lineheight{1.25}\smash{\begin{tabular}[t]{l}400 000\end{tabular}}}}%
    \put(0.90082951,0.05593334){\color[rgb]{0.00392157,0.00392157,0.00392157}\transparent{0.952941}\makebox(0,0)[lt]{\lineheight{1.25}\smash{\begin{tabular}[t]{l}500 000\end{tabular}}}}%
    \put(0.45313871,0.01082853){\color[rgb]{0.00392157,0.00392157,0.00392157}\transparent{0.952941}\makebox(0,0)[lt]{\lineheight{1.25}\smash{\begin{tabular}[t]{l}training steps\end{tabular}}}}%
  \end{picture}%
\endgroup%

%% file: conclusion_and_outlook.tex
This paper has presented a first attempt towards creating a link between
symbolic reward formulation based on informal descriptions of behaviour and
behaviour policy generation. The main novelty of our approach is a systematic
way to derive a reward function from an informal description of the desired
behaviour in three steps employing well-understood mathematical techniques: (i)
from an informal description of the behaviour, we have derived a formal
specification of the desired behaviour as a hybrid automaton; (ii) the  hybrid
automaton is an abstraction of the state space of the system, \ie a
restriction of its phase space; (iii) this restriction allowed to formulate a
better reward function for reinforcement learning. We have demonstrated that this kind
of input is useful in speeding up the learning process of achieving stable
bipedal locomotion; the key here is the effective search space reduction given
by the formal model.

It is worth pointing out that although we derive an effective reward function,
the main contribution of this paper is the way in which the combination of
symbolic description and reinforcement learning improves the overall result;
this should work with other reinforcement learning techniques and reward
functions as well, but to what degree remains to be investigated.
It further remains an open problem to figure out what is the right granularity to seek
in reward shaping. For example, it is interesting to study if splitting the
orthants into smaller hypercubes and coming up with more fine-grained rules will
bring additional improvements in the learning process. Furthermore, this formalization
naturally lends itself to applications in curriculum learning, where a coarse
grained initial formulation can be made more detailed when the agent reaches a
certain proficiency, which could potentially even more effectively guide the
learning agent. 

This kind of symbolic reward definition maybe be combined with existing inverse
reinforcement learning techniques to automatically derive reward specifications
for any given behaviour. For example, there is no principal reason why this
technique should not work for walking robots with higher degrees of freedom,
except that for these the formal model as a hybrid automaton may be not as
straightforward to formulate as for the compass walker. In future work, we want
to investigate how the present method can be applied to more complex humanoid
walking robots (such as \cite{esser:2020}) using the basics developed here
combined with mapping template models to whole body control.

Another aspect worth exploring is proving properties (such as safety properties)
about the controlled system. Here, the formalization as a hybrid automaton has
the advantage that powerful tool support exists~\cite{fulton2015,platzer2018}.
This can be used to show that learned behaviours satisfy given
properties~\cite{fulton2018}, so our approach can also be used for safety
verification of learned robot behaviours. This would be an enormous advantage
over existing approaches, where a safety guarantee for learned behaviours cannot
be given.